\documentclass[a4paper]{article}

\usepackage{INTERSPEECH2022}
\usepackage{paralist,multirow,url}
\usepackage{amssymb}% http://ctan.org/pkg/amssymb
\usepackage{pifont}% http://ctan.org/pkg/pifont
\newcommand{\cmark}{\ding{51}}%
\newcommand{\xmark}{\ding{55}}%

\title{Contrastive Learning for Improving ASR Robustness\\ in Spoken Language Understanding}
\name{Ya-Hsin Chang and Yun-Nung Chen}
\address{
  National Taiwan University, Taipei, Taiwan}
\email{r09922066@ntu.edu.tw\quad y.v.chen@ieee.org}

\begin{document}

\maketitle
\begin{abstract}
  Spoken language understanding (SLU) is an essential task for machines to understand human speech for better interactions. However, errors from the automatic speech recognizer (ASR) usually hurt the understanding performance.
  In reality, ASR systems may not be easy to adjust for the target scenarios.
  Therefore, this paper focuses on learning utterance representations that are robust to ASR errors using a contrastive objective, and further strengthens the generalization ability by combining supervised contrastive learning and self-distillation in model fine-tuning. Experiments on three benchmark datasets demonstrate the effectiveness of our proposed approach.\footnote{The codes are released in this github repository \url{https://github.com/MiuLab/SpokenCSE}.}
  
\end{abstract}

\section{Introduction}

Intelligent agents such as Apple Siri, Amazon Alexa, and Google Assistant are flourishing with recent advances in speech technology. The core element in these agents is spoken language understanding (SLU), which takes human speech input and extracts semantic information for various tasks, such as intent classification and slot filling. 
Existing SLU solutions can be categorized into two types: 1) pipeline (or cascade) approaches and 2) end-to-end approaches. Pipeline approaches first use automatic speech recognition (ASR) system to transcribe speech into text, followed by a natural language understanding (NLU) component for the target task, while end-to-end approaches~\cite{radfar2020end} apply a single model which can directly process speech signals and handle the understanding task without considering the text. 

Pipeline approaches take many benefits from using textual information: it would be easier to utilize additional resources including large-scale datasets and pre-trained models from the NLP community, relieve the burden of ASR development, and prevent privacy issues about using human voices.
However, pipeline methods usually suffer from error propagation -- when the ASR hypothesis is incorrect, the erroneous text can mislead the NLU model and hurt the performance. Although ASR systems today have already reached a low word error rate (WER) on data in a controlled environment, a real-world environment still leads to unsatisfied performance.
Researchers have explored remedies for ASR errors in two ways: either formulating the ASR error correction as a machine translation from erroneous ASR hypothesis to clean text~\cite{mani2020asr, wang2020asr, dutta2022error} or adapting the model through masked language modeling (MLM)~\cite{sundararaman2021phoneme, wang2022arobert, huang2019adapting}. Most prior approaches required additional \emph{speech-related} input features such as phoneme sequences~\cite{wang2020asr, dutta2022error, sundararaman2021phoneme}, lattice graph~\cite{huang2019adapting,zou2021associated,huang2020learning}, or N-best hypothesis~\cite{wang2021leveraging, zhu2021improving, ganesan2021n}. Such information may not be easily obtained due to the constraint of ASR systems.

With the rise of BERT~\cite{devlin2018bert}, pre-trained language models (PLMs) have been dominating the field of NLP. While it is straightforward to adopt PLM in pipeline SLU systems, PLMs are often trained on \emph{clean} text corpus and thus not resistant to ASR errors.
In order to learn the invariant representations between manual transcript and erroneous hypothesis, this paper proposes to utilize a contrastive objective to adapt PLM to ASR results with \emph{only textual information}.

Contrastive learning aims at pulling together the feature similarity of positive data pairs and pushing away negative data pairs~\cite{chen2020simple}. In computer vision, positive and negative samples are mostly derived from data augmentation, but due to the characteristics of discreteness in texts, the NLP community does not have a common strategy to create multiple views of a sentence to form positive samples.
Prior studies investigated different ways to construct positive pairs, such as back-translation~\cite{fang2020cert}, sampling from the same article~\cite{giorgi2020declutr}, or the pooled representation from different layers of a model~\cite{kim2021self}.
In spoken scenarios, we naturally take a manual transcript and its associated ASR hypothesis as a positive pair and maximize the similarity between their representations, because they come from the same audio signal.
Thus, the learned representations can be more error-robust through contrastive learning in \emph{pre-training}.

In addition, considering the heavily distorted data, we propose a supervised contrastive loss together with a self-distillation strategy during \emph{fine-tuning} in order to further strengthen the generalization capability.
Supervised contrastive learning is a variant of contrastive learning~\cite{khosla2020supervised}, where the positive/negative samples are data with the same/different labels, so the annotated target data is required.
Results on both vision~\cite{khosla2020supervised} and language~\cite{gunel2020supervised} showed improvement and robustness to input noises.
Self-distillation~\cite{zhang2019your}, or self-knowledge distillation, is a special form of knowledge distillation, where the teacher is the student itself. It simply minimizes Kullback-Leibler (KL) divergence with the model's previous prediction.
Without additional information from another model, self-distillation can still demonstrate the regularization effect and prevent over-confidence. We are the first to combine these two techniques for improving robustness in \emph{fine-tuning}.

The contributions of this work are four-fold:
\begin{compactitem}
\item We propose a novel contrastive objective for pre-training models robust to ASR errors. To our knowledge, we are the first to adopt such modeling techniques to improve robustness with only textual information.
\item We propose a novel fine-tuning framework combining supervised contrastive learning and self-distillation. 
\item The proposed method is flexible and can easily incorporate additional information such as phoneme or lattice.
\item Experiments on multiple benchmark datasets demonstrate that the proposed approach is capable of handling noisy text inputs and achieves significant improvement compared to other methods.
\end{compactitem}

\iffalse
\section{Related work}

Recent end-to-end approaches are trying to incorporate textual information for more resources. \cite{qian2021speech} further BERT with both speech and text embedding, and \cite{lai2021semi} jointly pre-trained an ASR model and a BERT in various settings. \cite{rao2021mean} adopted REINFORCE algorithm to directly optimize speech and text training at the same time. 

Similar ideas are also implemented in a supervised fashion. \cite{khosla2020supervised} showed the effectiveness of supervised contrastive training and the robustness to input noise in ImageNet. In natural language problems, \cite{gunel2020supervised} and \cite{zhang2021few} adopted this approach and reached the same conclusion. 
\fi

\section{Methodology}
Our proposed method consists of three elements: (1) A self-supervised contrastive objective for pre-training, (2) supervised contrastive learning, and (3) self-distillation in fine-tuning.

\begin{figure}[t!]
  \centering
  \includegraphics[width=\linewidth]{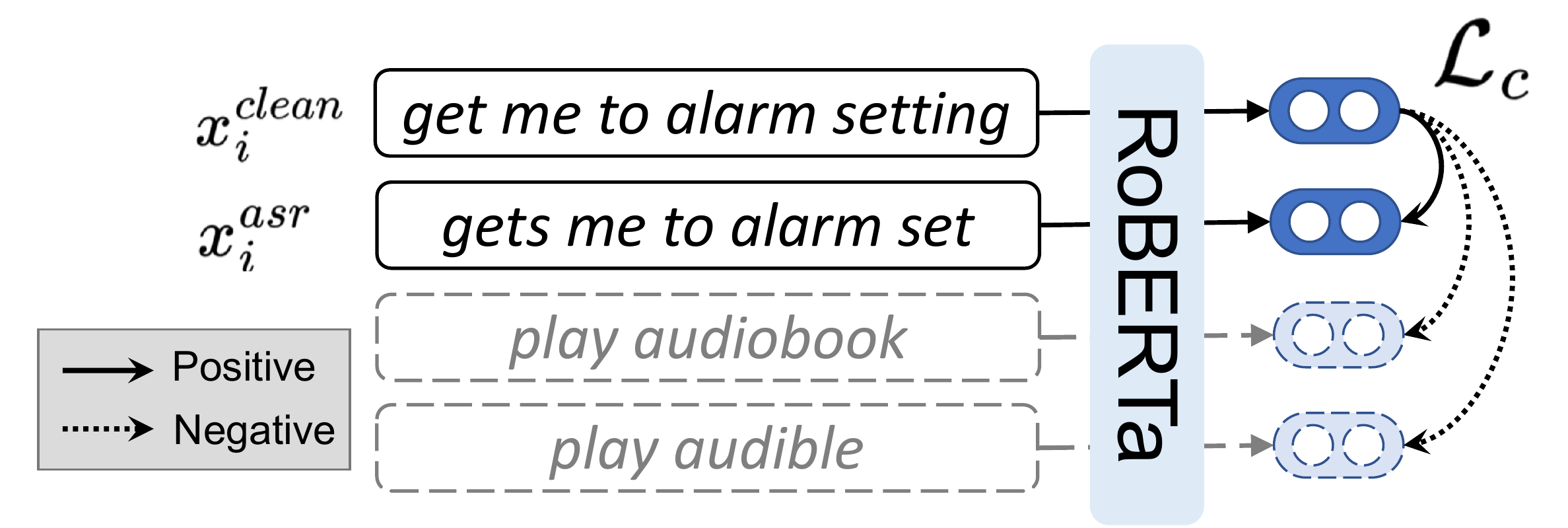}
  \vspace{-4mm}
  \caption{Pre-training: contrastive learning with the paired ASR noisy transcripts. A positive pair consists of clean data and ASR result from the same audio.}
  \label{fig:pretrain}
  \vspace{-5mm}
\end{figure}

\subsection{Self-supvervised contrastive learning}
Contrastive learning aims at helping our model distinguish the features invariant to input transformations, including data augmentations and corruptions.
To handle ASR errors, we propose to adopt contrastive learning for learning sentence representations invariant to misrecognition.
As shown in Figure \ref{fig:pretrain}, a pre-trained RoBERTa~\cite{liu2019roberta} is continually trained on spoken language corpus by utilizing the paired clean and noisy sentences.
%we propose to use contrastive learning to equip PLM with ASR error awareness. We initialized a BERT model from the pretrained RoBERTa and continually pretrain on spoken language corpus.

Given a mini-batch of input data of $N$ pairs of texts $B=\{(x^\text{clean}_i, x^\text{asr}_i)\}_{i=1..N}$ representing the clean manual transcript and ASR hypothesis. 
We first apply the pre-trained BERT and take the last layer of [CLS] to obtain the representation for each sentence as $h=\text{BERT}(x)$.
Then we further adjust the sentence representations by the proposed self-supervised contrastive loss~\cite{chen2020simple, gao2021simcse}:
\begin{equation}
  \begin{split}
    \mathcal{L}_{c} & = -\frac{1}{2N} \sum_{(h, h^{+}) \in \mathbb{P}} \log \frac{e^{s(h, h^{+})/\tau_{c}}} {\sum^B_{h' \neq h} e^{s(h, h') / \tau_{c}}} \\
    & = -\mathbb{E}_\mathbb{P} \Big[s(h, h^{+})/\tau_{c}\Big] + \mathbb{E}\Big[\log\big(  {\sum^B_{h' \neq h} e^{s(h, h') / \tau_{c}}} \big) \Big],
  \end{split}
  \label{eq1}
\end{equation}
where $\mathbb{P}$ is composed of $2N$ positive pairs of either $(h^\text{clean}_i, h^\text{asr}_i)$ or $(h^\text{asr}_i, h^\text{clean}_i)$, and \(s(\cdot,\cdot)\) is a cosine similarity function. 
The process is illustrated in Figure~\ref{fig:pretrain}.

Among two terms in the second line of (\ref{eq1}), the first term improves the \emph{alignment} between positive pairs with robustness to noise, and the second term promotes \emph{uniformity} in representation space by pushing away features of unrelated samples \cite{wang2020understanding}.
Both are known as good characteristics of representations and improve generalization.

To prevent catastrophic forgetting of the PLM, we keep the MLM objective in the pre-training process. Additionally, the prior work revealed the advantage of adaptive pre-training on the same domain of the downstream data \cite{gururangan2020don}, so the final proposed pre-training loss $\mathcal{L}_{pt}$ is the weighted sum of a contrastive loss $\mathcal{L}_c$ and an MLM loss $\mathcal{L}_{mlm}$:

\begin{equation}
  \mathcal{L}_{pt} = \mathcal{L}_c + \lambda_{mlm} \cdot \mathcal{L}_{mlm},
  \label{eq2}
\end{equation}
where $\lambda_{mlm}$ is a weight to maintain the model's ability of predicting the masked tokens.

\begin{figure*}[th]
  \centering
  \includegraphics[width=.98\textwidth]{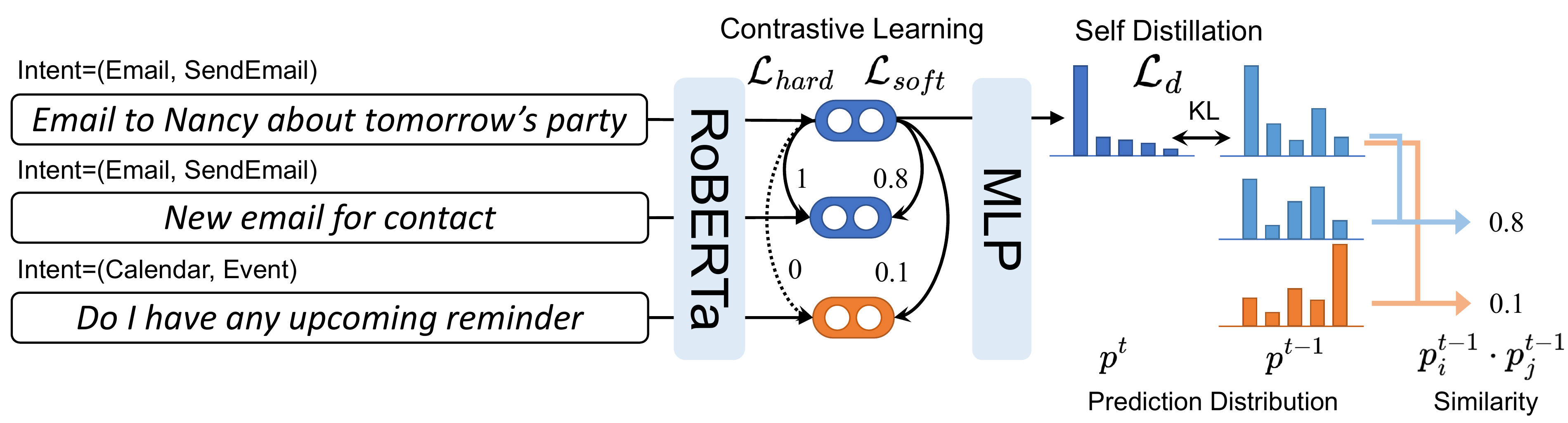}
  \vspace{-2.5mm}
  \caption{Fine-tuning: supervised contrastive learning with self-distillation}
  \label{fig:finetune}
  \vspace{-3mm}
\end{figure*}

\subsection{Supervised contrastive learning}
%In addition to apply contrastive learning in pre-training, it can also work at fine-tuning by using the label information.
Supervised contrastive learning at fine-tuning takes data of the same label as positive samples and pulls their embeddings closer together~\cite{khosla2020supervised}. In the end, the representations from the same label form a clustering effect and discriminate over different labels by creating margins between them.
This objective is similar to the widely-used triplet loss~\cite{weinberger2009distance}, but it can generalize to more than one positive and negative sample and is empirically shown to improve performance and resistance to input noises~\cite{gunel2020supervised}.
We propose to adopt a supervised contrastive loss $\mathcal{L}_{hard}$ to allow the learned representations aligned with their hard labels as illustrated in Figure~\ref{fig:finetune}:
%In our framework, our fine-tuning loss combines a supervised contrastive loss and a cross-entropy loss in the original fine-tuning stage:
\begin{equation}
  \begin{split}
    \mathcal{L}_{hard} = -\frac{1}{N} \cdot \sum^N_{i} \sum^N_{j\neq i} 1_{y_i=y_j} \log \frac{ e^{s(h_i, h_j)/\tau_{sc}}}{\sum^N_{k\neq i} e^{s(h_i, h_k) / \tau_{sc}}}.
  \end{split}
  \label{eq3}
\end{equation}
%\begin{equation}
%    L_{finetune} = L_{ce} + \lambda_{sc} L_{sc}
%    \label{eq4}
%\end{equation}

\subsection{Self-distillation}
Due to ASR errors, some input sentences may no longer retain the semantics of their labels, or shift to some fluent context of another class. For example, when \emph{on} is transcribed into \emph{off} in an IoT control command, the intent can be the exact opposite. Therefore, in order to reduce the impact of label noises in the training set, we propose a self-distillation method.

Self-distillation minimizes KL divergence between the current prediction and the previous one~\cite{yun2020regularizing, mobahi2020self}, which regularizes the model and eliminates label noise at the same time. We denote $p^{t}_i=P(y_i\mid x_i, t)$ as the probability distribution of data $x_i$ predicted by the model at the $t$-th epoch, and its loss function is formulated as: 
\begin{equation}
    \mathcal{L}_d = \frac{1}{N}\sum^N_i KL_{\tau_{d}}( p^{t-1}_i \| p^{t}_i ),
    \label{eq5}
\end{equation}
and $p^0_i$ is a one-hot vector of the label $y_i$.
This procedure is illustrated in the right part of Figure~\ref{fig:finetune}.

\subsection{Self-distilled soft contrastive learning}
To relieve the effect of noisy labels in supervised contrastive learning, we add a supplement loss similar to (\ref{eq3}) by contrasting the soft label calculated from the previous prediction:

\begin{equation}
  \begin{split}
    \mathcal{L}_{soft} = -\frac{1}{N} \sum^N_{i} \sum^N_{j\neq i}(p^{t-1}_i \cdot p^{t-1}_j) \log \frac{ e^{s(h_i, h_j)/\tau_{sc}}}{\sum^N_{k\neq i} e^{s(h_i, h_k) / \tau_{sc}}},
  \end{split}
  %\nonumber
  \label{eq6}
\end{equation}

This soft target strategy is also investigated in recent self-supervised contrastive learning studies such as ReSSL~\cite{zheng2021ressl} and SCE~\cite{denize2021similarity}, where the soft target comes from the similarity between their samples. 
In the framework shown in Figure~\ref{fig:finetune}, our finalized fine-tuning loss $\mathcal{L}_{ft}$ composes of four parts: 1) a cross entropy loss in the original fine-tuning stage $\mathcal{L}_{ce}$, 2) two contrastive learning losses (hard $\mathcal{L}_{hard}$ and soft $\mathcal{L}_{soft}$) and 3) a self-distillation loss $\mathcal{L}_{d}$ shown as below:
\begin{equation}
    \mathcal{L}_{ft} = \mathcal{L}_{ce} + \lambda_{d} \mathcal{L}_{d} + \lambda_{sc} \big( \mathcal{L}_{hard} +  \lambda_{d} \mathcal{L}_{soft} \big). 
    \label{eq7}
\end{equation}

\section{Experiments}

\begin{table}[th]
  \caption{Dataset statistics. SLURP test is sub-sampled.}
  \label{tab:datasets}
  \vspace{-2mm}
  \centering
  \begin{tabular}{ l r r r r}
    \toprule
    \multicolumn{1}{c}{\textbf{Dataset}} & 
        \multicolumn{1}{c}{\textbf{\#Class}} &
        \multicolumn{1}{c}{\textbf{Avg. Length}} &
        \multicolumn{1}{c}{\textbf{Train}} & 
        \multicolumn{1}{c}{\textbf{Test}} \\
    \midrule
    SLURP & \(18\times46\) & 6.93 & 50,628 & 10,992\\
    ATIS & 22 & 11.14 & 4,978 & 893  \\
    TREC6 & 6 & 8.89 & 5,452 & 500 \\
   \bottomrule
  \end{tabular}
  \vspace{-2mm}
\end{table}

\subsection{Datasets}
Three benchmark datasets are used for evaluating our model: SLURP~\cite{bastianelli2020slurp}, and two synthesized datasets ATIS and TREC6 from Phoneme-BERT ~\cite{sundararaman2021phoneme}. The statistics are shown in Table \ref{tab:datasets}. 

SLURP is a challenging SLU dataset with various domains, speakers, and recording settings. This paper only focuses on intent detection, where an intent is a (scenario, action) pair, and there are 18 scenarios and 46 actions in total, and the joint accuracy is used as the evaluation metric (both scenario and action are correct). 
We use two off-the-shelf ASR systems to obtain ASR hypothesis from the provided audio: Google Web API and wav2vec 2.0~\cite{baevski2020wav2vec}.\footnote{We use \emph{facebook/wav2vec2-large-960h} trained on LibriSpeech provided on HuggingFace~\cite{wolf2019huggingface}.}
The median word error rate (WER) is 25\% by Google and 60\% by wav2vec, implying the difficulty of performing SLU tasks using ASR hypothesis due to diverse accented speakers and noisy environments in this dataset.
Through manual inspection, we find some noisy or incorrect labels in SLURP shown in Table \ref{tab:labelnoise}, so we sub-sample a test set to ensure its quality for reliable evaluation.\footnote{We only keep samples where an ensemble of 5 RoBERTa models trained on manual transcripts gives the agreed prediction of their labels.}
Note that the training set may still contain label noises, so the model's robustness to label noises can be still validated in our experiments.

ATIS and TREC6 are two benchmark datasets for flight reservation and question classification respectively. We use the synthesized text released by Phoneme-BERT~\cite{sundararaman2021phoneme}, where the data is synthesized via a TTS model and later transcribed by ASR. Only a subset of data within a certain WER range is kept, and the reported average WER is 29.11\% for ATIS and 32.03\% for TREC6. We report accuracy as the evaluation metric.
%Phoneme-BErRT also targets on SLU without audio information, but it requires a language-model independent phoneme sequence generator. 
%Since we are not able to reproduce the generator, we adopt two of their datasets ATIS (flight reservation) and TREC6 (question classification) with phoneme sequences to draw a fair comparison. The data is synthesized by first converting original texts in the datasets via a text-to-speech model, then deriving the hypothesis from ASR and phoneme sequence from another generator. 

\begin{table}[t!]
  \caption{Label noises in SLURP (mislabeled and ambiguous). Model prediction is generated by a fine-tuned RoBERTa. 
  %Some data is mislabeled, and some can be described in multiple reasonable combinations of scenario and action.
  }
  \label{tab:labelnoise}
  \vspace{-2mm}
  \centering
  \begin{tabular}{ l c c}
    \toprule
    \multicolumn{1}{c}{\textbf{Transcripts}} & 
        \multicolumn{1}{c}{\textbf{Label}} &
        \multicolumn{1}{c}{\textbf{Prediction}} \\
    \midrule
\it hello how is your day & (general, \underline{quirky}) & (general, greet) \\
\it let's play music hits & (play, \underline{radio}) & (play, music) \\
\it what date is it & (\underline{calendar}, query) & (datetime, query)  \\
   \bottomrule
  \end{tabular}
  \vspace{-5mm}
\end{table}

\begin{table}[t]
  \caption{Results on three datasets. Phoneme-BERT$^\dagger$ additionally uses phoneme sequences generated by public tool-kit.}
  \label{tab:all}
  \vspace{-2mm}
  \centering
  \begin{tabular}{ l c c c}
    \toprule
    \textbf{Model} & \bf SLURP & \textbf{ATIS} & \textbf{TREC6} \\
    \midrule
    RoBERTa & 83.97 & 94.53 & 84.08\\
    Phoneme-BERT$^\dagger$ & 83.78 & 94.83 & 85.96 \\
    %Phoneme-BERT$^\ddagger$ & \cmark~~~~\cmark& \textbf{96.73} & 85.68   \\
    SimCSE & 84.47 & 94.07 & 84.92 \\
    Proposed (pre-train only) & 84.51 & 95.02 & 85.20 \\
    Proposed (pre-train + fine-tune) & \bf 85.26 & \bf 95.10 & \textbf{86.36} \\
   \bottomrule
  \end{tabular}
  \vspace{-5mm}
\end{table}

\subsection{Experimental setting}

%\subsection{Baselines}
We compare our model with three baselines: 
\begin{compactitem}
\item RoBERTa: a RoBERTa-base model directly fine-tuned on the target training data.
\item Phoneme-BERT~\cite{sundararaman2021phoneme}: a RoBERTa-base model further pre-trained on extra corpus with phoneme information. The phoneme sequences are tokenized by a RoBERTa tokenizer with a new token type embedding trained from scratch. 
Phoneme sequences are generated from the ASR hypothesis via a python toolkit\footnote{\url{https://github.com/bootphon/phonemizer}} since we do not have access to the phoneme decoder model.
\item SimCSE~\cite{gao2021simcse}: a state-of-the-art sentence embedding method using contrastive learning. We create positive pairs from two passes of the same ASR hypothesis for calculating $\mathcal{L}_c$ in (\ref{eq1}) so that we can better clarify the improvement from learning through manual transcripts in our proposed method.
%To clarify the improvement from learning information in manual transcripts, we implement SimCSE by modifying equation \ref{eq1} to create the positive pairs from two different passes of the same ASR hypothesis. The MLM objective remains the same.
\end{compactitem}
%Comparing with Phoneme-BERT indicates if our proposed method can better handle SLU using only textual information.

We pre-train the model for 10K steps with a batch size 128 on each task data, and fine-tune the model for 10 epochs with early stopping and a batch size 256.
%Without additional corpus, We apply pretraining on task data for each dataset. We pretrain the model for 10k steps with batch size 128, and finetune the model on target task for 10 epochs with early stopping and batch size 256. 
In SLURP, two separate classification heads are trained for scenario and action with shared BERT embeddings. 
%The hyperparameters across all experiments are listed in Table \ref{tab:hypers}.
We grid search over hyperparameters by metrics of validation set and find that the model is not sensitive, and the final setting is that the mask ratio of MLM is 0.15, $\tau_{c}= 0.2$, $\lambda_{mlm}=1$, $\tau_{sc}=0.2$, $\lambda_{sc}=0.1$, $\tau_{d}=5$, $\lambda_{d}=10$.
The reported scores are averaged over 5 runs.

\begin{table*}[t]
  \caption{Result on SLURP. WER intervals are separated by quartiles. The reported metric is joint accuracy. Phoneme-BERT uses input text with additional toolkit-generated phoneme sequences.}
  \label{tab:slurp}
  \vspace{-3mm}
  \centering
  \setlength\tabcolsep{4pt} % default value: 6pt
  \begin{tabular}{|cc|cccc|c|cccc|c|}
    \hline
    \multirow{3}{*}{\textbf{Pre-training}} & \multirow{3}{*}{\textbf{Fine-tuning}} & \multicolumn{5}{c|}{\textbf{SLURP WER Interval (Google)}} & \multicolumn{5}{c|}{\textbf{SLURP WER Interval (wav2vec)}} \\
    \cline{3-7}\cline{8-12}
    {} & {} & clean & low & medium & high & \multirow{2}{*}{all} & low & medium & high & severe & \multirow{2}{*}{all}\\
    {} & {} & =0 & (0,0.16] & (0.16~0.4] & $>$0.4 & {} & [0,0.25] & (0.25,0.5] & (0.5,0.83] & $>$0.83 & {} \\
    \hline
    RoBERTa & Direct & \bf 95.69 & 92.41 & 85.89 & 56.71 & 83.97 & 92.44 & 80.49 & 62.06 & 34.40 & 68.05 \\
    Phoneme-BERT & Direct & 94.97 & 92.34 & 85.87 & 57.20 & 83.78 & 91.40 & 79.71 & 62.60 & \textbf{36.56} & 68.24 \\
    % Phoneme-BERT($\dagger$) & Direct & 94.97 & 92.34 & 85.87 & 57.20 & 83.78 & 91.68 & 80.83 & 63.07 & 38.04 & 69.05 \\
    SimCSE & Direct & 95.55 & 93.47 & \bf 86.82 & 57.59 & 84.47 & 92.04 & 81.75 & 62.98 & 34.25 & 68.48 \\
    Proposed & Direct & 95.54 & \bf 93.86 & 86.68 & \bf 57.72 & \bf 84.51 & \bf 93.02 & \bf 82.56 & \bf  64.22 & 36.02 & \bf 69.66 \\
    \hline
    RoBERTa & Proposed & \textbf{96.59} & 94.27 & 86.70 & 57.24 & 84.87 & 93.56 & 81.59 & 63.29 & 34.61 & 68.98 \\
    Phoneme-BERT & Proposed & 95.61 & 93.42 & 86.87 & 57.50 & 84.48 & 92.12 & 81.46 & 61.48 & 33.64 & 67.89 \\
    SimCSE & Proposed & 96.57 & \textbf{94.54} & 87.39 & 58.01 & 85.25 & 92.62 & 80.86 & 62.13 & 33.25 & 67.94 \\
    Proposed & Proposed & 96.08 & 94.41 & \textbf{87.63} & \textbf{58.72} & \textbf{85.26} & \textbf{93.76} & \textbf{83.43} & \textbf{65.31} & \bf 35.83 & \bf 70.31 \\
  \hline
  \end{tabular}
  \vspace{-3mm}
\end{table*}

\iffalse
\begin{table}[th]
  \caption{Hyperparameters.}
  \label{tab:hypers}
  \centering
  \begin{tabular}{ r r}
    \toprule
    \multicolumn{1}{c}{\textbf{Hyperparameter}} & 
        \multicolumn{1}{c}{\textbf{Value}} \\
    \midrule
    MLM mask ratio & 0.15 \\
    \(\tau_{c}\) & 0.2 \\
    \(\lambda_{mlm}\) & 1 \\
    \(\tau_{c}\) & 0.2 \\
    \(\lambda_{c}\) & 0.1 \\
    \(\tau_{d}\) & 5 \\
    \(\lambda_{d}\) & 10 \\   
  \bottomrule
  \end{tabular}
\end{table}
\fi

\subsection{Evaluation results}

The evaluation performance is presented in Table~\ref{tab:all}, where three baselines only focus on pre-training for representation learning, so we additionally show the results only using our proposed pre-training method.
%It can be found that our proposed pre-approach significantly outperforms all baselines for all benchmark datasets.
The results demonstrate that the proposed contrastive learning in pre-training is useful for handling ASR noises and achieves better performance in most cases.
Phoneme-BERT 
%performs slightly better than the proposed pre-training method, but it 
requires additional speech information from phoneme sequences to boost the performance and thus less effective on SLURP, while our model only takes word-level information.
Moreover, the proposed fine-tuning framework with contrastive learning and self-distillation further models the uncertainty and improves the performance.
In summary, our proposed method is demonstrated effective for SLU and outperforms all baselines on three benchmark datasets.

\subsection{Analysis of different noise levels}
To better investigate the impact of different noise levels,  we separate the test set of SLURP into 4 groups according to their WER and show the performance in Table~\ref{tab:slurp}.
From the upper part of Table~\ref{tab:slurp}, our proposed pre-training procedure consistently outperforms all baselines for most cases except clean and severe noise levels.
When the input is clean, treating SLU as normal NLU is better, so the original RoBERTa performs best.
SimCSE pre-training achieves similar performance with relatively lower WER, but cannot generalize to noisier inputs when we use wav2vec transcripts.
Because our proposed self-supervised contrastive learning method allows the model to learn the invariant features by utilizing the relationship between ASR and manual transcripts, the learned spoken representations can be robust even to much noisy inputs.
%with Google ASR hypothesis; however, when training on noisy Wav2vec hypothesis, the improvement becomes insignificant. Our proposed method helps the model learn the invariant features with clean text information, while SimCSE is not able to realize the possible recovery from noisy inputs. 
%Therefore, in most cases, our proposed pre-training method that utilizes self-supervised contrastive learning is demonstrated useful for SLU and outperforms prior strong baselines.
From the lower part of Table~\ref{tab:slurp}, our proposed fine-tuning procedure can further boost the performance regardless of pre-trained methods, showing the effectiveness of our framework combining supervised contrastive learning and self-distillation.
%Phoneme-BERT gives little improvement with phoneme sequences generated from erroneous texts; they may contain too much noise to provide useful information. 

\iffalse
\begin{table}[t]
  \caption{Results on ATIS and TREC6. %Phoneme-BERT$^\ddagger$ takes both text and phoneme sequences as inputs.
  }
  \label{tab:atis-trec6}
  \vspace{-1.5mm}
  \centering
  \begin{tabular}{ l c c c c}
    \toprule
    \multirow{2}{*}{\textbf{Model}} & \bf Phoneme &
        \multirow{2}{*}{\textbf{ATIS}} &
        \multirow{2}{*}{\textbf{TREC6}} \\
        & \bf \small Pre-train~~Fine-tune & & \\
    \midrule
    RoBERTa & \xmark~~~~\xmark & 94.53 & 84.08\\
    Phoneme-BERT & \cmark~~~~\xmark & 94.85 & 85.56 \\
    Phoneme-BERT$^\ddagger$ & \cmark~~~~\cmark& \textbf{96.73} & 85.68   \\
    SimCSE & \xmark~~~~\xmark& 94.07 & 84.92 \\
    Proposed & \xmark~~~~\xmark & 95.10 & \textbf{86.36} \\
   \bottomrule
  \end{tabular}
  \vspace{-3mm}
\end{table}

\subsection{Analysis of Phoneme Information}
%Considering that phoneme information may provide additional signal for recovering misrecognitions, 
Table \ref{tab:atis-trec6} presents the results of two Phoneme-BERT datasets, ATIS and TREC6.
Pre-training on large-scale language model-independent phonemes does have a strong influence on performance and remains effective without the phoneme sequence in the downstream task. Nevertheless, our proposed method outperforms Phoneme-BERT when no additional information is available. Again, SimCSE pre-training on noisy data gives little improvement.
\fi

\begin{table}[t!]
  \caption{Ablation study of different losses (\%).}
  \label{tab:ablation}
  \vspace{-3mm}
  \centering
  \begin{tabular}{ l l c c c c}
    \toprule
        $\mathcal{L}_{pt}$ &
        $\mathcal{L}_{ft}$ &
       \textbf{SLURP} &
        \textbf{ATIS} &
        \textbf{TREC6} \\
    \midrule
    full & full & \bf 85.26 & \bf 95.10 & \bf 86.36 \\
    no $\mathcal{L}_{mlm}$ & full & 84.83 & 93.75 & 85.32 \\
    no $\mathcal{L}_c$ & full & 85.15 & 95.00 & 85.52 \\
    %\midrule
    full & no $\mathcal{L}_{hard}+\mathcal{L}_{soft}$ & 85.14 & 94.83 & 86.08\\
    full & no $\mathcal{L}_{d}+\mathcal{L}_{soft}$ & 84.77 & 94.75 & 85.60 \\
    full & no $\mathcal{L}_{soft}$ & 84.81 & 94.65 & 86.20\\
   \bottomrule
  \end{tabular}
  \vspace{-3mm}
\end{table}

%\section{Discussion}
\subsection{Ablation study}
Because the proposed method is composed of multiple elements, we conduct an ablation study to further investigate the effect of each loss in Table \ref{tab:ablation}.
%, where the and we verify all components contribute positively.
The results show that each proposed method (for pre-training and fine-tuning) contributes to the performance positively.
It is obvious that MLM in pre-training and self-distillation in fine-tuning play an important role, because MLM pre-training is crucial to adapting the PLM to noisy inputs, and self-distillation prevents the model from overfitting the mismatched text and label pairs.
%as the accuracy on the training set often surpasses 95\%.
Moreover, self-distillation also reduces the impact of label noises in SLURP and thus brings more contribution.

\begin{table}[t]
  \caption{Results on SLURP with parallel data in different ways. %\texttt{Correction} indicates that we train a ASR correction model with a small decoder and fine-tune the encoder. 
  %\texttt{Gold}, \texttt{Hypo}, and \texttt{Mixed} represent manual transcript, ASR hypothesis or both when fine-tuning.
  }
  \vspace{-3mm}
  \label{tab:parallel}
  \centering
  \begin{tabular}{ ll c}
    \toprule
    \textbf{Pre-train Method} &
    \textbf{Fine-tune Data} &
    \multicolumn{1}{c}{\textbf{Accuracy}}  \\
    \midrule
    ASR Correction  & ASR & 83.96 \\
    Proposed  & ASR & 85.26 \\
    %\midrule
    %RoBERTa & Manual + ASR & 84.03 \\
    Proposed & Manual & 84.82 \\
    Proposed & Manual + ASR & \bf 85.64 \\
   \bottomrule
  \end{tabular}
  \vspace{-4mm}
\end{table}

\subsection{Analysis of exploiting manual transcripts}
In our proposed procedure, we pre-train on the paired ASR and manual transcripts via contrastive learning.
Such information can be exploited in other ways; 
for example, we can pre-train a sequence-to-sequence model for correcting ASR hypothesis to its manual transcript, and then the encoder can be used in the fine-tuning stage.
The upper part of Table~\ref{tab:parallel} shows that the proposed approach can better utilize the paired signal and achieves better performance.
One possible reason is that the limited paired data is insufficient for correcting ASR results, while our proposed contrastive learning does not focus on fine-grained text correction but on representation distance for great effectiveness and better efficiency. 

Furthermore, our experiments assume that only ASR results are available during fine-tuning for better practice, because the original audio signal may not be manually transcribed for privacy issues.
If both manual and ASR transcripts are available in the downstream task, we can also fine-tune the model on both types of data.
The lower part of Table \ref{tab:parallel} shows that additionally utilizing manual transcripts in fine-tuning is beneficial, but taking ASR transcripts is necessary to align with the target inference scenario.
Future work can consider how to deal with the scenarios when only clean sentences are available. 
Also, even without SLU data, our proposed method can still utilize any speech corpus with off-the-shelf ASR in our pre-training stage.
Hence, our future work is to validate if other available large speech corpus can further improve the robustness of ASR through our proposed method.

%Finetuning with only manual transcript hurts the performance because of data shift in the testing stage while using both can provide improvement. It would be an interesting direction to develop methods to finetune with clean text only, but it is beyond the scope of this study.
%for a wider application
\section{Conclusions}

This work introduces a novel contrastive objective for learning ASR-robust representations and utilizes supervised contrastive learning and self-distillation to better handle the uncertainty and prevent overfitting.
To our knowledge, we are not only the first to utilize contrastive learning for modeling the invariant features between manual and ASR transcripts but also the first to combine self-distillation with supervised contrastive learning for better handling uncertainty.
Experiments on three benchmark datasets demonstrate the effectiveness of the proposed framework, showing the great potential of bridging the gap of understanding performance between clean and noisy inputs.
%and usefulness of each proposed component . Detailed experiments identify the contribution of each component and show the proposed method is efficient in exploiting a parallel corpus of manual transcripts and ASR hypothesis. In the future, we hope to combine our approach with other techniques and information to bridge the gap of NLU performance between using clean text and ASR hypothesis.

\iffalse
\section{Acknowledgements}

The ISCA Board would like to thank the organizing committees of the past INTERSPEECH conferences for their help and for kindly providing the template files. \\
Note to authors: Authors should not use logos in the acknowledgement section; rather authors should acknowledge corporations by naming them only.
\fi

\bibliographystyle{IEEEtran}
\bibliography{mybib}

% \begin{thebibliography}{9}
% \bibitem[1]{Davis80-COP}
%   S.\ B.\ Davis and P.\ Mermelstein,
%   ``Comparison of parametric representation for monosyllabic word recognition in continuously spoken sentences,''
%   \textit{IEEE Transactions on Acoustics, Speech and Signal Processing}, vol.~28, no.~4, pp.~357--366, 1980.
% \bibitem[2]{Rabiner89-ATO}
%   L.\ R.\ Rabiner,
%   ``A tutorial on hidden Markov models and selected applications in speech recognition,''
%   \textit{Proceedings of the IEEE}, vol.~77, no.~2, pp.~257-286, 1989.
% \bibitem[3]{Hastie09-TEO}
%   T.\ Hastie, R.\ Tibshirani, and J.\ Friedman,
%   \textit{The Elements of Statistical Learning -- Data Mining, Inference, and Prediction}.
%   New York: Springer, 2009.
% \bibitem[4]{YourName17-XXX}
%   F.\ Lastname1, F.\ Lastname2, and F.\ Lastname3,
%   ``Title of your INTERSPEECH 2022 publication,''
%   in \textit{Interspeech 2022 -- 23\textsuperscript{rd} Annual Conference of the International Speech Communication Association, September 18-22, Incheon, Korea, Proceedings, Proceedings}, 2022, pp.~100--104.
% \end{thebibliography}

\end{document}